\documentclass{article}
\usepackage{multirow}

\usepackage[main,final]{neurips_2026}
\makeatletter
\renewcommand{\@notice}{}
\makeatother
\usepackage[utf8]{inputenc} 
\usepackage[T1]{fontenc}    
\usepackage{hyperref}       
\usepackage{url}            
\usepackage{booktabs}       
\usepackage{amsfonts}       
\usepackage{nicefrac}       
\usepackage{microtype}      
\usepackage{xcolor}         
\usepackage{graphicx}

\title{ExtrAnom: A VAD Benchmark Emphasizing the Need for Women’s Safety in Society}

\author{
Sangeeta\\
Indian Institute of Technology Bhubaneswar\\
Bhubaneswar, Odisha 752050, India\\
\texttt{a24cs09005@iitbbs.ac.in}
\And
Maddikuntla Sai Prajwal\\
Indian Institute of Technology Bhubaneswar\\
Bhubaneswar, Odisha 752050, India\\
\texttt{22cs01045@iitbbs.ac.in}
\And
Debi Prosad Dogra\\
Indian Institute of Technology Bhubaneswar\\
Bhubaneswar, Odisha 752050, India
\And
Kamalakar Vijay Thakare\\
Artificial Intelligence and Robotics Institute,\\
Korea Institute of Science and Technology\\
Seoul 02792, Republic of Korea\\
\texttt{tkv15@iitbbs.ac.in}
\AND
Hyungjoo Jung\\
Artificial Intelligence and Robotics Institute,\\
Korea Institute of Science and Technology\\
Seoul 02792, Republic of Korea
\And
Ig-Jae Kim\\
Artificial Intelligence and Robotics Institute,\\
Korea Institute of Science and Technology\\
Seoul 02792, Republic of Korea
\And
Heeseung Choi\\
Yonsei-KIST Convergence Research Institute,\\
Yonsei University\\
Seoul 03722, Republic of Korea
}

\begin{document}

\maketitle

\begin{abstract}
Women's safety and security are paramount for a modern society. Often, crimes scenes get recorded through low-resolution CCTV cameras limiting the efficiency of video anomaly detection (VAD) models. Despite substantial progress in VAD research, women-centric anomalies are still underrepresented in datasets as well as in models. Existing datasets primarily cover well-lit, high-resolution and close-shot videos that are inadequate to  tackle critical anomalies such as chain snatching, stalking, inappropriate touch, and other subtle forms of crime against women. To address this, we present a new benchmark, referred to as ExtrAnom. It contains 1001 videos (both anomalies and normal) with four textual annotations; one human-generated and three LLM-generated. The videos are arranged in 5 different categories of crimes. The dataset comprises low-light (8\%), low-resolution (13\%), long-shot (15\%), and  daytime (64\%) anomaly videos. It includes stalking  (3.9\%), chain snatching (17.6\%), kidnapping (7.3\%), assassinations (2.3\%), harassment (18.9\%), and normal (50\%) videos. It is possible to perform cross-modal and VLM-based validations using ExtrAnom. We have benchmarked it against popular unimodal and multi-modal VAD datasets (e.g., XD-Violence, UCF-Crime, and UCA) and SOTA methods. Experiments reveal that existing datasets are insufficient to deal with women-centric anomalies. We believe ExtrAnom can fill this critical gap in VAD research.
\end{abstract}

\section{Introduction}
Video Anomaly Detection (VAD) plays a critical role in identifying unusual events and behaviors within public safety systems. Recent advancements in artificial intelligence and computer vision have facilitated effective anomaly detection across various domains, including traffic~\cite{trafficanom2,trafficanom}, industrial settings~\cite{industrial2,industrial1}, and incidents of public-place violence~\cite{weaklysupervised,gods,UCF-crime,UCA}. Furthermore, the emergence of vision-language models~\cite{holmesvau} has enhanced VAD by utilizing textual information to identify fine-grained anomalies and improve interpretability.

\begin{figure}[t]
    \centering
    \includegraphics[width=0.75\linewidth]{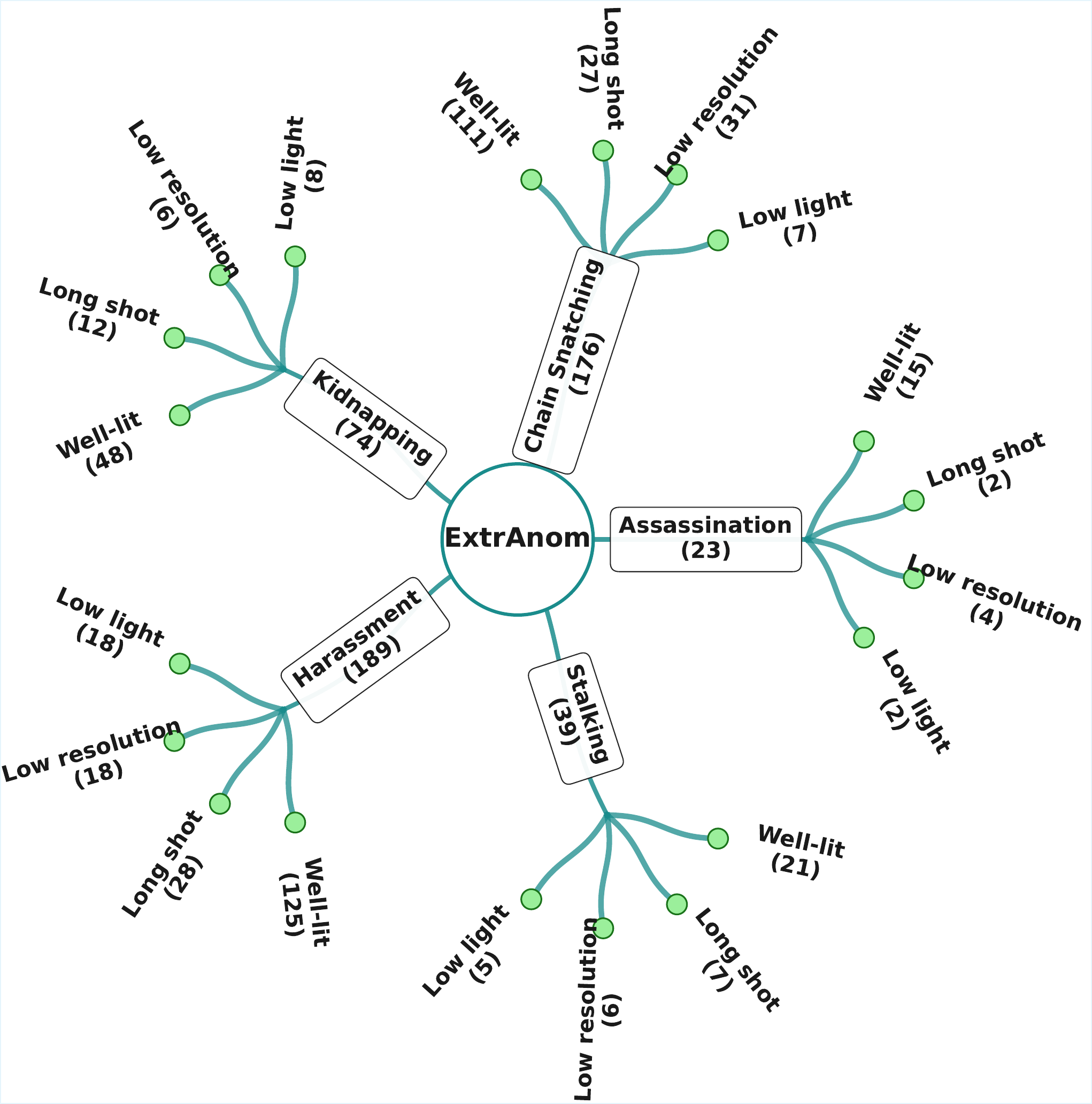}
    \caption{Visual representation of video proportions among five different anomalous categories. This radial tree shows that, within each category, the numbers of low-lighting, long-shot-view, low-resolution, and well-lit/daylight videos are consistent.}
    \label{fig:data}
\end{figure}
Women's safety remains a critical societal concern, with crimes such as harassment, stalking, chain snatching, and kidnapping occurring frequently in public spaces. According to the UN Women and UNODC report~\cite{un}, around 50,000 women and girls were killed in 2024 due to abuse, while the WHO reports that nearly one in three women experiences violence during her lifetime~\cite{WHO}. Among these crimes, \textbf{stalking} is particularly challenging to identify because it closely resembles normal human interactions, making its early detection crucial for preventing more severe crimes. Although several VAD datasets have been proposed~\cite{UCSD,subway,cuhk,UCF-crime,UCA,xdviolence,RareAnom}, they primarily focus on general or event-specific anomalies. Existing datasets include only limited women-related incidents and largely overlook important categories such as stalking, chain snatching, and inappropriate touch, thereby failing to capture the diversity of crimes against women.

Due to a lack of adequate data, existing models fail to detect women-related anomalies. For example, there is only one video about chain snatching on UCF-Crime~\cite{UCF-crime}, and the recent-most anomaly descriptor Holmes-VAU~\cite{holmesvau} describes the event as: {\it``The anomaly event involves a man in a white shirt and black pants riding a motorcycle and then stopping to talk to a woman in a blue dress, which is suspicious and unusual behavior that deviates from the normal traffic flow and pedestrian activity depicted in the rest of the video"}. The method summarizes the video as a normal event. But the actual incident was: {\it ``Two men dressed in dark clothing were riding a motorcycle when they slowed down beside a woman wearing a blue saree. One of the men suddenly snatched her chain, and the pair immediately sped away from the scene.
"}. Due to a lack of women-centric data, the model leads to an incorrect conclusion, such as normal interaction, accident, or robbery of a vehicle. Therefore, arranging a dataset suitable for women's safety can be a timely intervention. The present work fills this gap by proposing {\bf ExtrAnom:} a dataset containing videos of crimes against women. The number of videos for each category is shown in Fig.~\ref{fig:data}.

Descriptions generated using VLM Holmes-VAU~\cite{holmesvau} trained on UCF-Crime are erroneous, e.g. $62.45\%$ (on ExtrAnom) and $37.80\%$ (on UCF-Crime~\cite{UCF-crime}) incorrect descriptions. Around $65.78\%$ of ExtrAnom videos and $15.48\%$ of UCF-Crime~\cite{UCF-crime} videos have been misclassified due to the lack of women-centric data including inappropriate touch, stalking, chain snatching, and assassination. XD-Violence~\cite{xdviolence} and other available datasets contain videos recorded under controlled environment with good lighting and higher resolution, limiting their applicability to real-world surveillance. 
\begin{figure}[ht]
  \centering  
  \includegraphics[width=0.7\linewidth]{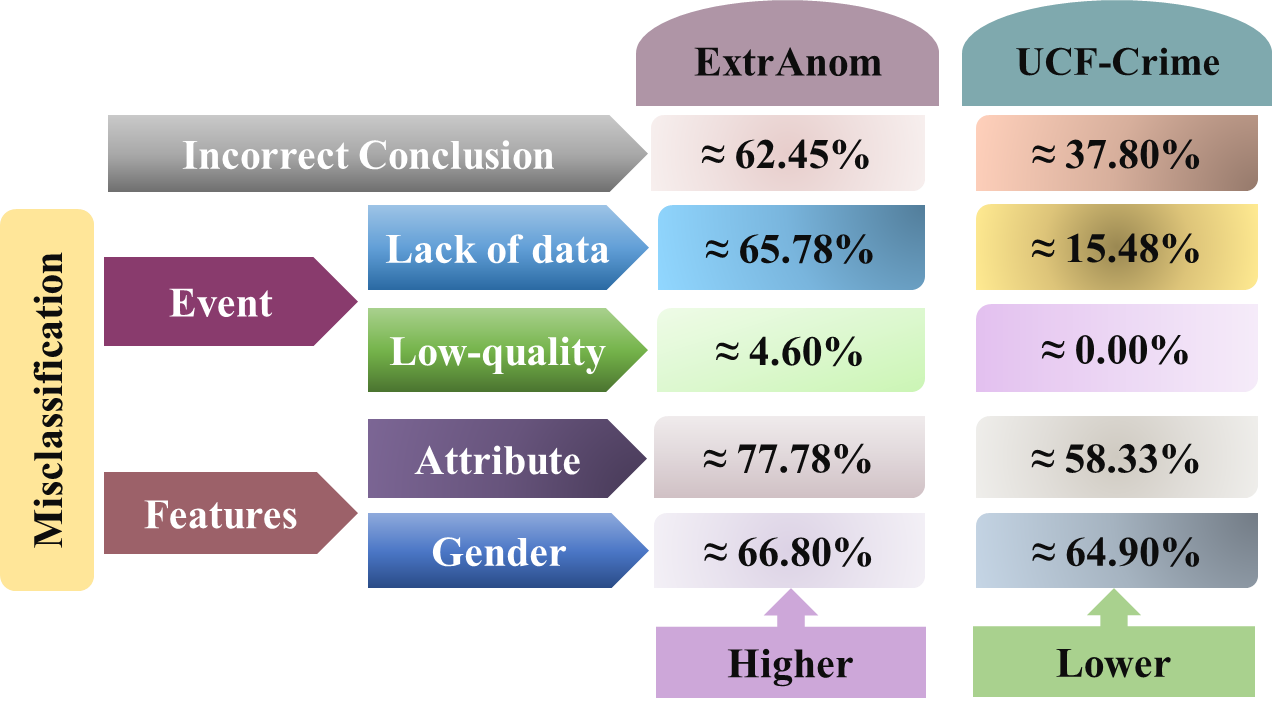}
  \caption{Comparison of errors produced by the pretrained Holmes-VAU~\cite{holmesvau} model on UCF-Crime~\cite{UCF-crime} and ExtrAnom datasets.}
  \label{fig:comparision}
  \end{figure}
Though VLMs perform  better than vision-based models, however, they require large textual annotations, which are not included in RareAnom, XD-Violence, Subway, ADOC, CCTVFight, and StreetScene~\cite{RareAnom,xdviolence,subway,ADOC,cctvfight,streetscene} datasets. Therefore, Holmes-VAU~\cite{holmesvau}, InternVL3~\cite{intern}, and QwenVL3~\cite{qwen} under-perform. Holmes-VAU~\cite{holmesvau} has misclassified the gender for $66.8\%$ of ExtrAnom videos and $64.9\%$ of UCF-Crime~\cite{UCF-crime} videos that related to extreme anomalies against women, as shown in Fig.~\ref{fig:comparision}. Moreover, both VLM-based~\cite{holmesvau,intern} and vision-based models~\cite{gods,RareAnom} misclassify dangerous objects such as guns and knives. Fig.~\ref{fig:comparision} shows that Holmes-VAU~\cite{holmesvau} has misclassified $77.78\%$ of dangerous objects in the ExtrAnom dataset and $58.33\%$ in the UCF-Crime~\cite{UCF-crime} dataset.

\begin{figure}[h]
    \centering
    \includegraphics[width=0.9\linewidth]{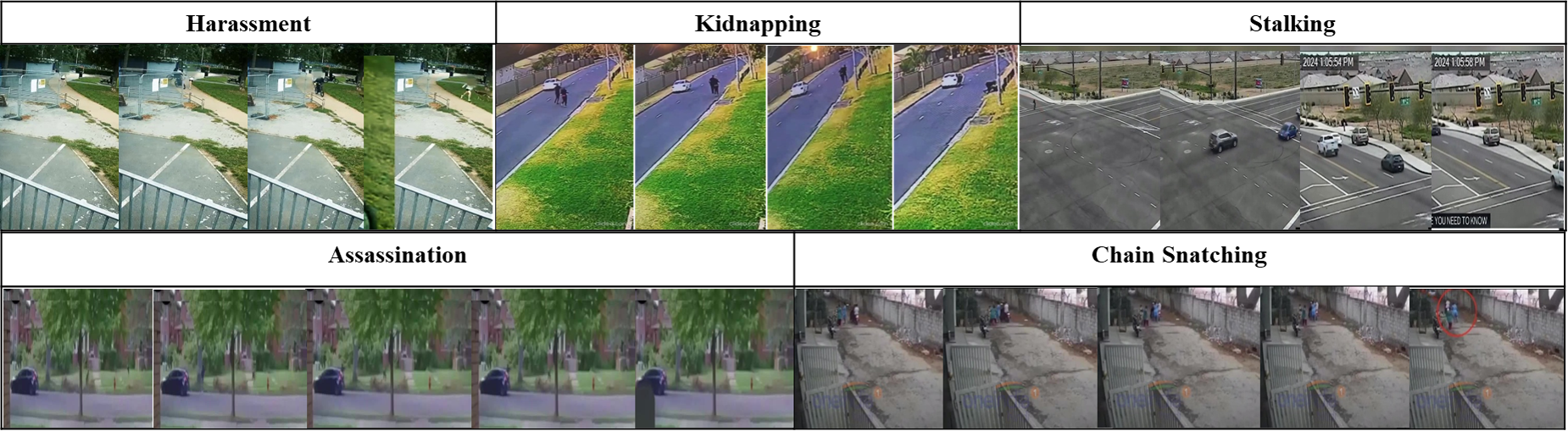}
    \caption{A few examples of the videos recorded using long shot views in each category. }    
    \label{fig:cameradistance}
\end{figure}

To address these issues, we present ExtrAnom, a dataset containing extreme-level anomalies against women. It contains 1001 videos collected from publicly available repositories recorded in an uncontrolled environment. Some of these videos were captured under daylight conditions, and some were in extremely low-light conditions, with poor resolution, as depicted in Fig.~\ref{fig:lowlight}. Some videos were captured in long shot with different types of sensors (CCTV, handheld cameras, mobile phones, etc), as depicted in Fig.~\ref{fig:cameradistance}. We have collected these videos from multiple sources, including Twitter, YouTube, websites, etc. For each video, multiple textual descriptions have been created. We have conducted extensive experimental validations to establish the challenging features and relevance of the dataset. The following are the technical contributions of the paper:
\begin{figure}[ht]
    \centering
    \includegraphics[width=0.9\linewidth]{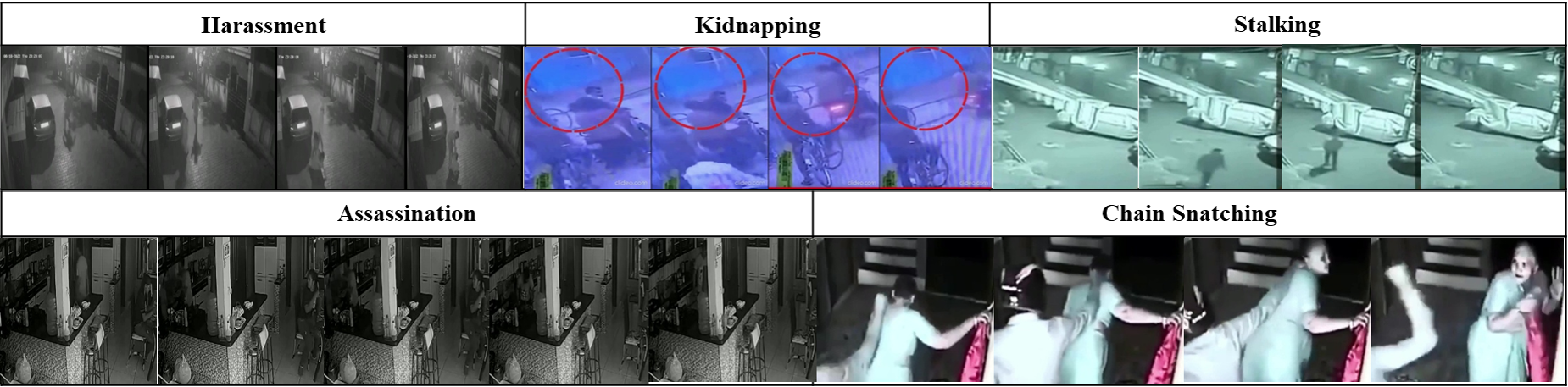}
    \caption{A few examples of low-light and low-resolution videos of each category. }    
    \label{fig:lowlight}
\end{figure}
 \begin{itemize}
     \item We release a new real-world VAD dataset focusing on women-centric crimes, referred to as ExtrAnom. It contains a total of 1001 videos, 500 normal and 501 anomalous, categorized into 5 types of anomalies. This is the first dataset that includes events related to inappropriate touch, stalking, and chain snatching involving women.
  
     \item We release textual annotations of the ExtrAnom videos, including ground truths generated by human annotators and 3 LLM-generated descriptions with prompts. 
     
     \item We present extensive experimental analysis  using SOTA anomaly detection methods to explain the advantages and drawbacks of these methods. 
 \end{itemize}
The rest of the paper is organized as follows. We first review the related work. Next, we describe the ExtrAnom dataset and its statistics. We then present the benchmarking and evaluation results. Finally, we conclude the paper and discuss future directions.

\section{Related Work}
\label{sec:related}
\textbf{Video Anomaly Detection Datasets:} The VAD datasets are publicly available for anomaly detection and evaluating the performance of video anomaly detection models. Subway Entrance~\cite{subway} and Subway Exit~\cite{subway} contain one large video, and anomaly types are walking in wrong direction, no payment, jumping, or squeezing, etc. UCSD-Ped1~\cite{UCSD} and UCSD-Ped2~\cite{UCSD} contain soft anomalies such as bike or cart movements. CUHK Avenue~\cite{cuhk} contains 37 types of anomalies, including run, throw, and new objects. The street scene~\cite{streetscene} dataset contains two-way urban street scenes, containing 17 anomaly types, including a biker outside the street, a person sitting on the beach, dogs on the sidewalk, etc. CCTV-Fights~\cite{cctvfight} contains only fight videos, and ADOC~\cite{ADOC} contains 25 anomaly types, including riding a bike, birds flying, and cat/dog movements. UBnormal~\cite{ubnormal} contains 22 anomalies, including running, falling, fighting etc. All the above datasets contain medium-scale and soft anomalies. UCF-crime~\cite{UCF-crime}, UCA~\cite{UCA}, RareAnom~\cite{RareAnom}, and XD violence~\cite{xdviolence} are large-scale, widely used datasets in the field of anomaly detection. These datasets contain frequently occurring anomalies, including person falls, abuse, fighting etc. UCF-crime\cite{UCF-crime} and UCA\cite{UCA} are single-source datasets, captured using CCTV. UCA\cite{UCA} is an extended form of UCF-Crime\cite{UCF-crime}, which contains textual descriptions of events for VLMs. XD-violence~\cite{xdviolence} is captured from various sources, including CCTV, Movies, animation, and performed by actors. RareAnom~\cite{RareAnom} is a multi-source, multi-view dataset that contains 17 classes, including women abuse and kidnapping. However, most of the datasets are collected under a controlled environment, contain high-resolution, intense lighting,  soft anomaly videos, and do not contain a detailed description of the event to train multi-modal LLMs. Limiting its capability to distinguish from extreme anomalies in real-world scenarios with a textual description.

\noindent\textbf{Video Anomaly Detection Methods:} The VAD models are widely used in the field of road safety, industrial safety, and to detect any kind of anomaly. Vision-based models are classified into semi-supervised~\cite{semi1,semi2}, weakly-supervised ~\cite{UCF-crime,rtfm,mist} and unsupervised ~\cite{gods,RareAnom,dyannet}. Weakly supervised models ~\cite{UCF-crime,rtfm,mist} use both normal and anomalous videos with video-level annotation to train a model that differentiates between abnormal behavior to normal behavior. Unsupervised models require only normal videos to train the model and do not rely on annotated data. Recently, multi-modality methods involving video and audio~\cite{videoaudio} or video and language~\cite{intern,holmesvau,videollava,llama} have also been introduced in the field of anomaly detection. Holmes-VAU~\cite{holmesvau} performs well in generating clip-level descriptions, but struggles in event-level and video-level descriptions.
\section{The ExtrAnom Dataset}
\label{sec:dataset}
 We introduce the ExtrAnom dataset focused on extreme violence against women. In this section,  we include (i) the importance of the dataset, (ii) data creation and annotation, (iii) comparisons with existing datasets, and (iv) analysis.

\subsection{Why is it Important?}
ExtrAnom dataset addresses a few critical gaps in VAD research. (i) Existing VAD datasets~\cite{UCF-crime,xdviolence,RareAnom} have only a small number of videos of kidnapping and harassment involving women. (ii) No dataset covers incidents like inappropriate touch, chain snatching, or stalking. (iii) Datasets such as XD-Violence~\cite{xdviolence} largely contain high-resolution, well-lit, and staged videos. (iv) UCF-Crime~\cite{UCF-crime} dataset videos are close-shot in nature and lack the discriminating features that are necessary to understand distant anomalies related to women. (v) Inappropriate touch, which is part of the harassment category, usually happens at public and crowded places, and no dataset addresses this category. (vi) A multidimensional reference set is unavailable in other datasets. Thus, we propose ExtrAnom with a wide variety of interpretive viewpoints and descriptive emphasis to address the aforementioned shortcomings.

\subsection{Collection and Annotation Protocols}
\noindent\textbf{Collection and Sourcing.}
ExtrAnom was sourced from publicly available CCTV footage, news channels, and social media platforms (e.g., Twitter and YouTube) using predefined search phrases for each anomaly category (see supplemental material). Videos were only included if they clearly depicted the desired event and the major subject. The dataset includes real-world videos collected by CCTV, mobile phones, and handheld cameras under various settings such as low resolution, poor illumination, and long-shot views, making anomaly recognition tough. Duplicate videos were manually eliminated, resulting in a single clip for each incident. Normal videos do not contain any abnormal scene. Moreover, videos with unrelated anomalies such as traffic accidents, were eliminated to prevent label leakage.

\begin{figure}[ht]
 \centering    \includegraphics[width=\linewidth]{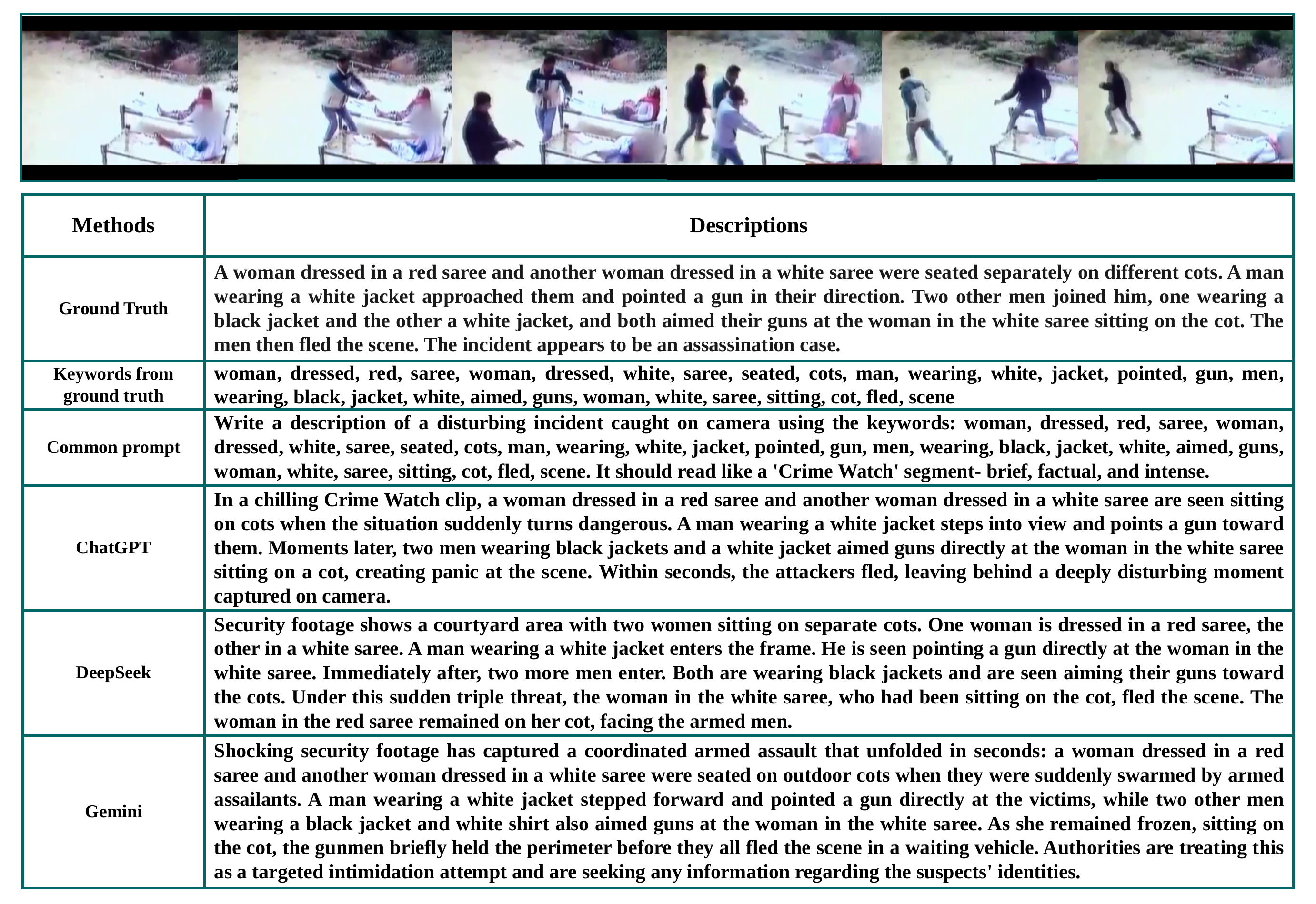}
    \caption{\textbf{Textual Descriptions:} The figure shows descriptions of an event generated by a human and 3 LLMs such as ChatGPT\cite{gpt}, DeepSeek\cite{deepseek}, and Gemini~\cite{gemini}. Human annotators write ground truth and keywords. To generate the descriptions using LLMs, we have used keywords and a prompt. }
    \label{fig:description}
\end{figure}

\noindent\textbf{Textual Annotations:} Textual annotation aims to provide detailed video descriptions for VLMs. Each video includes one human-authored ground-truth description and three keyword-conditioned LLM descriptions (ChatGPT, DeepSeek, and Gemini). Each video was annotated over multiple stages to ensure label quality. Three annotators created the original descriptions and category labels; two additional annotators independently examined and revised them; and an independent annotator went over all videos to ensure uniformity. Disagreements discovered during the review were resolved by correction at the checking step. An example description of an event is shown in Figure~\ref{fig:description}.

\noindent\textbf{Enhancement of Textual Descriptions:}
We fed the LLMs with keywords (objects, actions, persons with attributes) to produce enhanced descriptions of the events. To quantitatively evaluate the quality, we computed the semantic similarity between the generated descriptions and the human-written ground truths using SBERT~\cite{sbert} embeddings. Table~\ref{tab:llm_fidelity} illustrates that LLMs such as ChatGPT~\cite{gpt}, DeepSeek~\cite{deepseek}, and Gemini~\cite{gemini} have added 51.6\%, 49.0\%, and 55.7\% more relevant words to the original descriptions, respectively. The keyword-guided prompting substantially improves the semantic similarity with the human-written ground truths. For instance, the similarity score of DeepSeek~\cite{deepseek} increases from $0.28$ to $0.72$, ChatGPT~\cite{gpt} from $0.46$ to $0.74$, and Gemini~\cite{gemini} from $0.35$ to $0.71$. These improvements indicate that the generated descriptions are more aligned with the visual contents. The resulting LLM-generated narratives are included in ExtrAnom as additional textual annotations, providing richer supervision for VLMs.


\begingroup

\setlength{\tabcolsep}{8pt} 
\renewcommand{\arraystretch}{1.18} 
\begin{table}[tbh]
\centering
\caption{LLM-assisted description quality before and after keyword-guided refinement. \emph{Important words} denotes the percentage increase in relevant words.}
\label{tab:llm_fidelity}
\resizebox{\textwidth}{!}{%
\begin{tabular}{lccccc}
\hline
\multicolumn{1}{c}{\multirow{2}{*}{\textbf{Source}}} &
  \multicolumn{2}{c}{\textbf{Avg. Words}} &
  \multirow{2}{*}{\textbf{Important Words}} &
  \multicolumn{2}{c}{\textbf{Similarity}} \\ \cline{2-3} \cline{5-6} 
\multicolumn{1}{c}{}     & \textbf{Before} & \textbf{After}         &                                             & \textbf{Before} & \textbf{After}         \\ \hline
Human (GT)               & 56.6            & \textcolor{blue}{56.6} & 56,640                                      & 1.00            & \textcolor{blue}{1.00} \\
DeepSeek~\cite{deepseek} & 34.5            & \textcolor{blue}{65.5} & 84,394 (\textcolor{blue}{$\uparrow$49.0\%}) & 0.28            & \textcolor{blue}{0.72} \\
ChatGPT~\cite{gpt}       & 83.4            & \textcolor{blue}{61.9} & 85,866 (\textcolor{blue}{$\uparrow$51.6\%}) & 0.46            & \textcolor{blue}{0.74} \\
Gemini~\cite{gemini}     & 47.7            & \textcolor{blue}{83.5} & 88,188 (\textcolor{blue}{$\uparrow$55.7\%}) & 0.35            & \textcolor{blue}{0.71} \\ \hline
\end{tabular}%
}
\end{table}

\endgroup

 \subsection{ExtrAnom vs Others}
Existing datasets~\cite{UCF-crime,xdviolence,RareAnom} mainly focus on generalized anomalies and contain only a limited number of women-related events, such as purse snatching, abuse, and kidnapping. They do not adequately cover diverse women-centric anomalies, including chain snatching, stalking, harassment (inappropriate public touching in crowded places), assassination, and kidnapping. Consequently, these datasets provide insufficient supervision for training VLMs to understand fine-grained women-centric interactions. Table~\ref{tab:comparison} compares existing datasets with ExtrAnom in terms of video diversity, anomaly categories, and key characteristics, highlighting the unique contributions of ExtrAnom to women-centric video anomaly detection.

\begin{table}[h]
\centering
\scriptsize
\setlength{\tabcolsep}{2pt}
\caption{Comparison of ExtrAnom with existing video anomaly detection datasets.}
\label{tab:comparison}
\resizebox{0.75 \textwidth}{!}{%
\begin{tabular}{lccccc}
\toprule
\textbf{Dataset} & \textbf{\#Vid} & \textbf{Cls} & \textbf{Women} & \textbf{Text} \\
                 &                &              & \textbf{Centric} & \textbf{Anno.} \\
\midrule
UCF-Crime~\cite{UCF-crime}          & 1900 & 13 & Partial & No \\
UCA~\cite{UCA}                      & 1854 & 13 & Partial & Yes \\
UBnormal~\cite{ubnormal}            & 543  & 22 & No      & No \\
ADOC~\cite{ADOC}                    & 1    & 25 & No      & No \\
Street Scene~\cite{streetscene}     & 81   & 17 & No      & No \\
CCTV-Fights~\cite{cctvfight}        & 1000 & 1  & No      & No \\
Subway Entrance~\cite{subway}       & 1    & 1  & No      & No \\
XD-Violence~\cite{xdviolence}       & 4754 & 6  & Partial & No \\
RareAnom~\cite{RareAnom}            & 2200 & 17 & Partial & No \\
UCSD-Ped1~\cite{UCSD}               & 70   & 5  & No      & No \\
UCSD-Ped2~\cite{UCSD}               & 28   & 5  & No      & No \\
CUHK Avenue~\cite{cuhk}             & 37   & 5  & No      & No \\
\midrule
\textbf{ExtrAnom (Ours)}            & \textbf{1001} & \textbf{5} & \textbf{Yes} & \textbf{Yes} \\
\bottomrule
\end{tabular}
}
\end{table}
Table~\ref{tab:womcomp} highlights the limited representation of women-centric anomaly categories in existing datasets and only covers harassment-related events. Important categories such as chain snatching, stalking, kidnapping, and assassination are either absent or underrepresented.
\begin{table}[h]
\centering

\setlength{\tabcolsep}{3pt}
\caption{Comparisons of women-centric anomaly categories in existing datasets. CS, ST, KD, HT, and AS denote Chain Snatching, Stalking, Kidnapping, Harassment (Inappropriate Touch), and Assassination, respectively. Values indicate the number of videos available for each category.}
\label{tab:womcomp}
\begin{tabular}{lccccc}
\toprule
\textbf{Dataset} & \textbf{CS} & \textbf{ST} & \textbf{KD} & \textbf{HT} & \textbf{AS} \\
\midrule
UCF-Crime~\cite{UCF-crime} 
& 1 & nil & nil & 47(nil) & nil \\
UCA~\cite{UCA} 
& 1 & nil & nil & 47(nil) & nil \\
XD-Violence~\cite{xdviolence} 
& nil & nil & nil & 64(nil) & nil \\
RareAnom~\cite{RareAnom} 
& nil & nil & 16 & 70(nil) & nil \\
\midrule
ExtrAnom(our) 
& 176 & 39 & 74 & 189(86) & 23 \\
\bottomrule
\end{tabular}
\end{table}

\subsection{Statistical Analysis}
Table~\ref{tab:analysis} provides statistics of ExtrAnom in five categories of women-centric anomalies. The dataset includes 501 anomalous videos with 235,371 frames and durations ranging from 6.1 s (Chain Snatching) to 28.1 s (Kidnapping), documenting both short- and long-duration anomalous incidents. Harassment (189 videos) and Chain Snatching (176 videos) provide a huge number of training samples, while Stalking, Kidnapping, and Assassination are relatively uncommon but crucial real-world scenarios. Videos are collected in demanding surveillance conditions, such as low-light (4.0-12.8\%), long-shot (8.7-17.9\%), and low-resolution (8.1-17.6\%) settings, which closely mirror practical CCTV surroundings. ExtrAnom's qualities make it a realistic and demanding benchmark for creating strong video anomaly detection models that may be applied to real-world women-centric surveillance applications.

\begin{table}[h]
\centering

\setlength{\tabcolsep}{3pt}
\caption{Statistics of ExtrAnom across anomaly categories. Avg. Dur. denotes average video duration (s); Low-light, Long-shot, and Low-resolution indicate the percentage of videos captured under challenging visual conditions.}
\label{tab:analysis}
\begin{tabular}{lcccccc}
\toprule
\textbf{Anomaly} & \textbf{\#Vid} & \textbf{\#Frames} & \textbf{Dur.} & \textbf{LL} & \textbf{LS} & \textbf{LR} \\
\midrule
Assassination   & 23  & 18422  & 26.7 & 8.7  & 8.7  & 17.4 \\
Kidnapping      & 74  & 60785  & 28.1 & 10.8 & 16.2 & 8.1 \\
Stalking        & 39  & 21142  & 19.1 & 12.8 & 17.9 & 15.5 \\
Harassment      & 189 & 104477 & 18.8 & 9.5  & 14.8 & 9.5 \\
Chain Snatching & 176 & 31545  & 6.1  & 4.0  & 15.3 & 17.6 \\
\bottomrule
\end{tabular}
\end{table}

\subsection{Privacy Consent}
The ExtrAnom videos have been collected from publicly available sources such as news channels, Twitter, and other online platforms avoiding stricter copyright compliance. The dataset will be available on request for research purposes only.   

\section{Benchmarking and Evaluation}
\label{sec:benchmarking}
We have evaluated ExtrAnom utilizing cutting-edge multi-modal LLMs and vision-based anomaly detection models. Existing models trained in UCF-Crime~\cite{UCF-crime} and XD-Violence~\cite{xdviolence} are directly assessed on ExtrAnom to examine their ability to recognize women-centric anomalies without task-specific adaptations. We then train SOTA anomaly detection models using ExtrAnom to assess the effectiveness of our proposed benchmark. The training set consists of 800 (400 abnormal and 400 normal) and the test set contains 201 (101 abnormal and 100 normal) videos. To ensure a fair comparison, all methods have been evaluated under similar experimental settings. 

\subsection{Performance of Multi-modal LLMs}

\begin{table}[t]
\centering
\scriptsize
\setlength{\tabcolsep}{2pt}
\caption{The table presents the performance
of different multi-modal LLMs in our dataset. We evaluate the quality of video-level descriptions generated by existing multi-modal LLMs and compare them with the ground truth.}
\label{tab:similarityscore}
\resizebox{0.8 \textwidth}{!}{%
\begin{tabular}{lccccc}
\toprule
\textbf{Method} & \textbf{BLEU} & \textbf{BERT} & \textbf{CIDEr} & \textbf{METEOR} & \textbf{ROUGH} \\
\midrule
Video-ChatGPT~\cite{videochatgpt}      & 0.0163 & 0.2627 & 0.2082 & 0.1818 & 0.2411 \\
Video-LLaMA~\cite{llama}               & 0.0138 & 0.1947 & 0.1639 & 0.1625 & 0.2069 \\
Video-LLaVA~\cite{videollava}          & 0.0059 & 0.1409 & 0.1159 & 0.1367 & 0.1374 \\
LLaVA-Next-Video~\cite{llavanext}      & 0.0056 & 0.0192 & 0.1271 & 0.0648 & 0.1120 \\
QwenVL3~\cite{qwen}                    & 0.0118 & 0.2233 & 0.0009 & 0.2376 & 0.2014 \\
InternVL3~\cite{intern}                & 0.0061 & 0.2100 & 0.0004 & 0.2116 & 0.1439 \\
Gemini~\cite{gemini}                   & 0.0197 & 0.2946 & 0.0866 & 0.2572 & 0.2739 \\
Holmes-VAU~\cite{holmesvau}            & 0.0201 & 0.1956 & 0.2397 & 0.1423 & 0.2243 \\
\midrule
\multicolumn{6}{c}{\textbf{Fine-tuned}} \\
\midrule
Holmes-VAU~\cite{holmesvau}
& \textcolor{blue}{0.0824$\uparrow$}
& \textcolor{blue}{0.3270$\uparrow$}
& \textcolor{blue}{0.3820$\uparrow$}
& \textcolor{blue}{0.1743$\uparrow$}
& \textcolor{blue}{0.3850$\uparrow$} \\
\midrule
\multicolumn{6}{c}{\textbf{Training-free VLM}} \\
\midrule
LAVAD~\cite{lavad}                     & 0.0058 & 0.1271 & 0.0720 & 0.1314 & 0.1327 \\
\bottomrule
\end{tabular}
}
\end{table}

\begin{table}[t]
\centering
\small
\setlength{\tabcolsep}{2.5pt}
\caption{Performance comparisons (AUC \%) of anomaly detection methods under zero-shot inference and train/test settings. RA, UCF, XD, and Ext. denote RareAnom, UCF-Crime, XD-Violence, and ExtrAnom, respectively.}
\label{tab:modeleval}
\resizebox{0.77\textwidth}{!}{%
\begin{tabular}{lcc|cccc}
\toprule
{\textbf{Method}} &
{\textbf{Feat.}} &
\multicolumn{1}{c|}{\textbf{Infer.}} &
\multicolumn{4}{c}{\textbf{Train + Test}} \\
\cmidrule(lr){3-3}\cmidrule(lr){4-7}
& & \textbf{Ext.} & \textbf{RA} & \textbf{UCF} & \textbf{XD} & \textbf{Ext.} \\
\midrule
MIL~\cite{UCF-crime}        & C3D & 49.62 & 60.17 & 77.92 & 73.20 & \textcolor{red}{55.63$\downarrow$} \\
BODS~\cite{gods}            & I3D & 39.25 & 55.46 & 68.26 & 57.32 & \textcolor{red}{49.52$\downarrow$} \\
GODS~\cite{gods}            & I3D & 40.63 & 57.60 & 70.46 & 61.56 & \textcolor{red}{53.72$\downarrow$} \\
RTFM~\cite{rtfm}            & I3D & 44.56 & --    & 84.30 & 77.81 & \textcolor{red}{65.39$\downarrow$} \\
MIST~\cite{mist}            & I3D & 41.60 & 66.77 & 82.30 & 77.92 & \textcolor{red}{57.25$\downarrow$} \\
RareAnom~\cite{RareAnom}    & I3D & 40.89 & 62.39 & 78.86 & 68.33 & \textcolor{red}{55.47$\downarrow$} \\
DyAnNet~\cite{dyannet}      & I3D & 53.25 & --    & 84.50 & --    & \textcolor{red}{67.53$\downarrow$} \\
CLIP-TSA~\cite{cliptsa}     & ViT & 51.00 & --    & 87.58 & 82.19 & \textcolor{red}{61.20$\downarrow$} \\
VadCLIP~\cite{vadclip}      & ViT & 50.60 & --    & 88.02 & --    & \textcolor{red}{70.89$\downarrow$} \\
\bottomrule
\end{tabular}
}
\end{table}

\begin{figure}[t]
    \centering
    \includegraphics[scale=0.50]{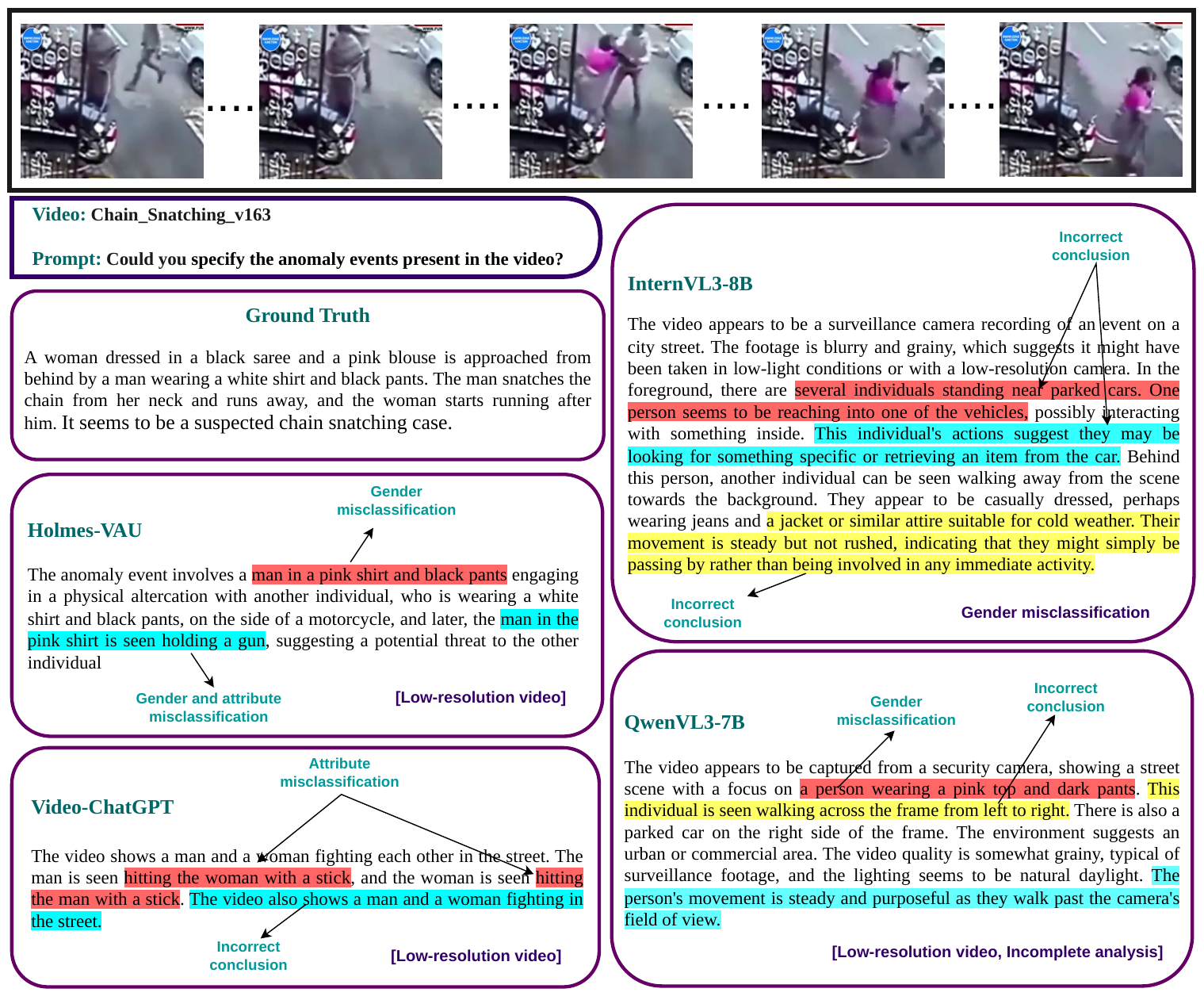}
    \caption{The figure shows how existing multi-modal LLMs misclassify attributes and gender, leading to incorrect conclusions and assumptions due to erroneous object detection. For example, Holmes-VAU and Video-ChatGPT incorrectly identify a man as holding a gun and a stick, respectively. }    
    \label{fig:similaritycheck}
\end{figure}

We use eight pretrained multi-modal LLMs ~\cite{videochatgpt,llama,videollava,llavanext,qwen,intern,gemini,holmesvau} trained on UCF-Crime~\cite{UCF-crime} and XD-Violence~\cite{xdviolence} to generate video-level descriptions of ExtrAnom. The generated descriptions are compared to the ground truth using BLEU~\cite{bleu}, BERT~\cite{bert}, CIDEr~\cite{cider}, METEOR~\cite{meteor}, and ROUGE~\cite{rouge}. According to Table~\ref{tab:similarityscore}, Video-LLaMA~\cite{llama}, Video-LLaVA~\cite{videollava}, and LLaVA-Next-Video~\cite{llavanext} have low similarity scores, while the remaining models have only moderate scores. We fine-tuned Holmes-VAU~\cite{holmesvau} on the training split, resulting in significant gains across all metrics. We also assess the training-free VLM LAVAD~\cite{lavad}, which performs similarly poorly. Overall, present multi-modal LLMs struggle to detect women-centric anomalies, which frequently mimic regular human interactions. 

We provide a descriptive analysis of different Multi-modal LLMs and ground truths. Figure~\ref{fig:similaritycheck} shows a sample description of chain snatching, and it shows that existing multi-modal LLMs generate inaccurate and incomplete descriptions. Even the most recent model, e.g., Holmes-VAU~\cite{holmesvau}, also suffers from gender and attribute misclassifications. In several cases, the model incorrectly classifies woman as man, detecting other objects as guns, etc. Due to the erroneous identification of a man holding a gun, the overall interpretation of the event can shift from a chain snatching incident to shooting. As a result, the conclusion of the model changes significantly. Similarly, the Video-ChatGPT~\cite{videochatgpt} model incorrectly describes the man and woman hitting each other with a stick, resulting in a change in interpretation of the event from a chain snatching incident to a physical fight. InternVL3-8B~\cite{intern} detects the presence of a vehicle and assumes that the person arrives using one of the vehicles. The model further infers cold weather conditions and fails to identify any unusual behavior in the video. Moreover, it is unable to determine the gender of the individual involved or wrongly estimates the number of participants in the event. As a result, InternVL3-8B~\cite{intern} classifies the video as normal, concluding that no anomalous activity occurred. This leads to an incorrect interpretation, highlighting the limitations of a model in gender classification and event understanding. Similarly, QwenVL3-7B~\cite{qwen} is unable to identify whether the person is a man or a woman and provides an incomplete description of the event, failing to capture the purpose of an individual. The examples illustrate that existing models frequently misidentify participants' genders, overlook subtle interactions, or infer unrelated activities. These errors suggest limitations in understanding context-dependent women-centric events rather than only recognizing general violent actions.

\subsection{Performance of Vision-based Models}
The effect of ExtrAnom has been evaluated using SOTA vision-based models and compared with three widely used anomaly datasets: RareAnom~\cite{RareAnom}, UCF-Crime~\cite{UCF-crime}, and XD-Violence~\cite{xdviolence}. Initially, the vision-based models have been trained on existing datasets and subsequently tested on ExtrAnom. Models have been trained on these existing datasets achieved only 39–53 AUC when directly transferred to ExtrAnom, despite substantially higher performance on their original benchmarks. This significant performance gap suggests that ExtrAnom introduces challenges not adequately represented in current anomaly datasets. Subsequently, the models have been trained and tested on ExtrAnom. After training on ExtrAnom, performance has improved from 49 to 70 AUC, indicating that the dataset offers complementary supervision beyond existing benchmarks. Table~\ref{tab:modeleval} presents the performance of unsupervised and weakly supervised models on ExtrAnom. Though performance increases after training on ExtrAnom, however, it still remains considerably lower as compared to UCF-Crime~\cite{UCF-crime}, XD-Violence~\cite{xdviolence}, and RareAnom~\cite{RareAnom}. This highlights the difficulty level presented in ExtrAnom.

\subsection{Why Women-Centric?}
\begin{figure}[t]
    \centering
    \includegraphics[scale=0.4]{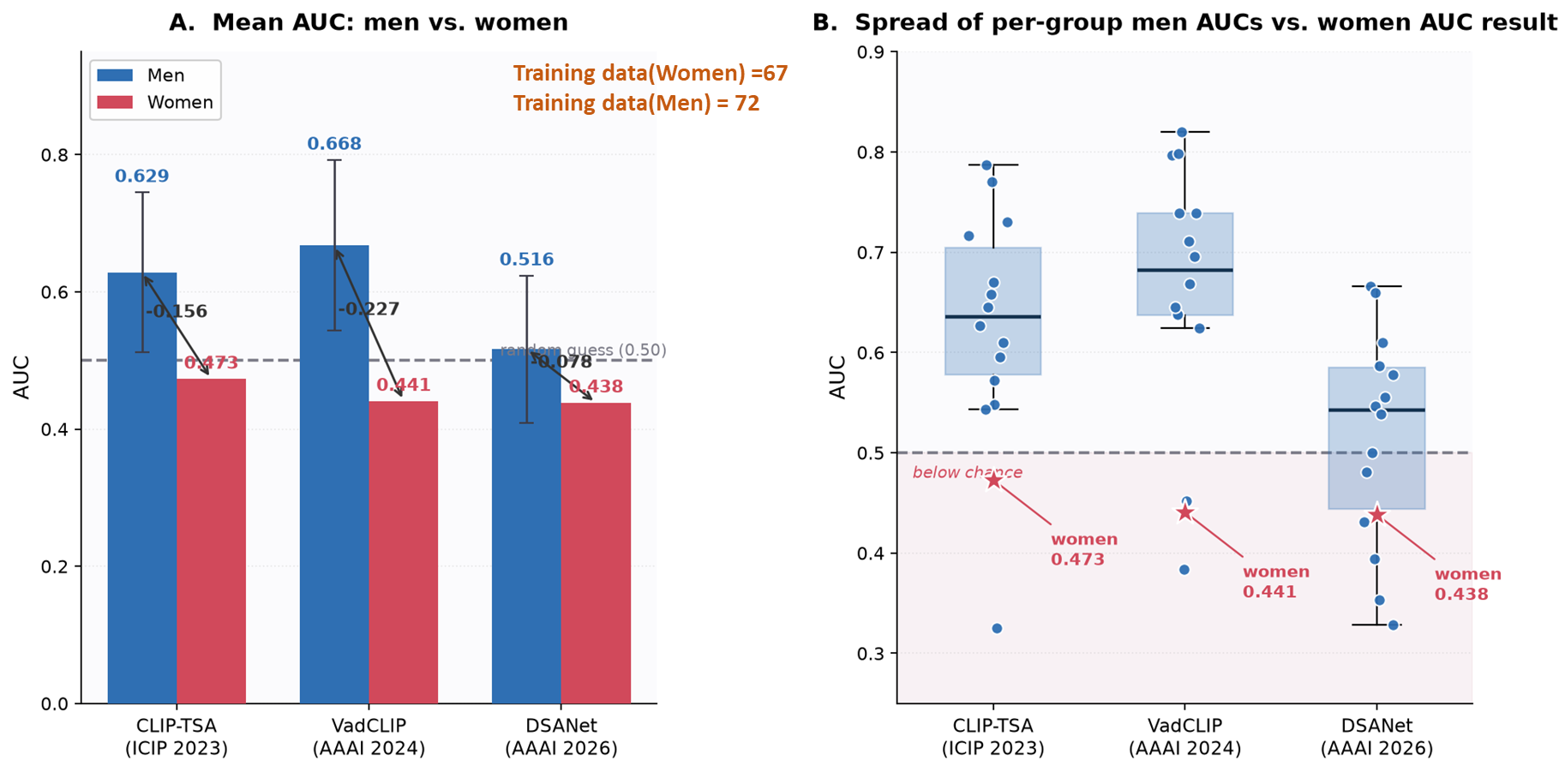}
    \caption{Comparisons of VAD performance on men- and women-centric anomalies. Existing models consistently achieve lower AUC on women-centric anomalies.}
    \label{fig:gender_auc}
\end{figure}

The current VAD datasets are highly imbalanced with respect to gender. For example, UCF-Crime contains 256 men-centric  and 67 women-centric videos, while XD-Violence contains 668 men-centric as compared to 64 women-centric videos. Such imbalance biases existing models toward men-centric anomalies. To ensure a fair comparison, we have evaluated all methods using a fair distribution of men and women videos. Even under such a balanced setting, the average AUC drops from $0.604$ (men-centric) to $0.451$ (women-centric) anomalies, with every women-centric result falls below the random-guess threshold of $0.50$, as shown in Fig.~\ref{fig:gender_auc}. Notably, these experiments have been conducted on UCF-Crime~\cite{UCF-crime}, where the videos are predominantly well-lit, high-resolution, and captured from close-shot views. Despite these favorable conditions, SOTA VAD models struggle to recognize women-centric anomalies, suggesting that the performance gap arises from the lack of gender-diverse training data rather than poor video quality. These observations motivate the need for a dedicated benchmark for evaluating women-centric video anomalies.

\begin{figure}[tbh]
    \centering
    \includegraphics[scale=0.305]{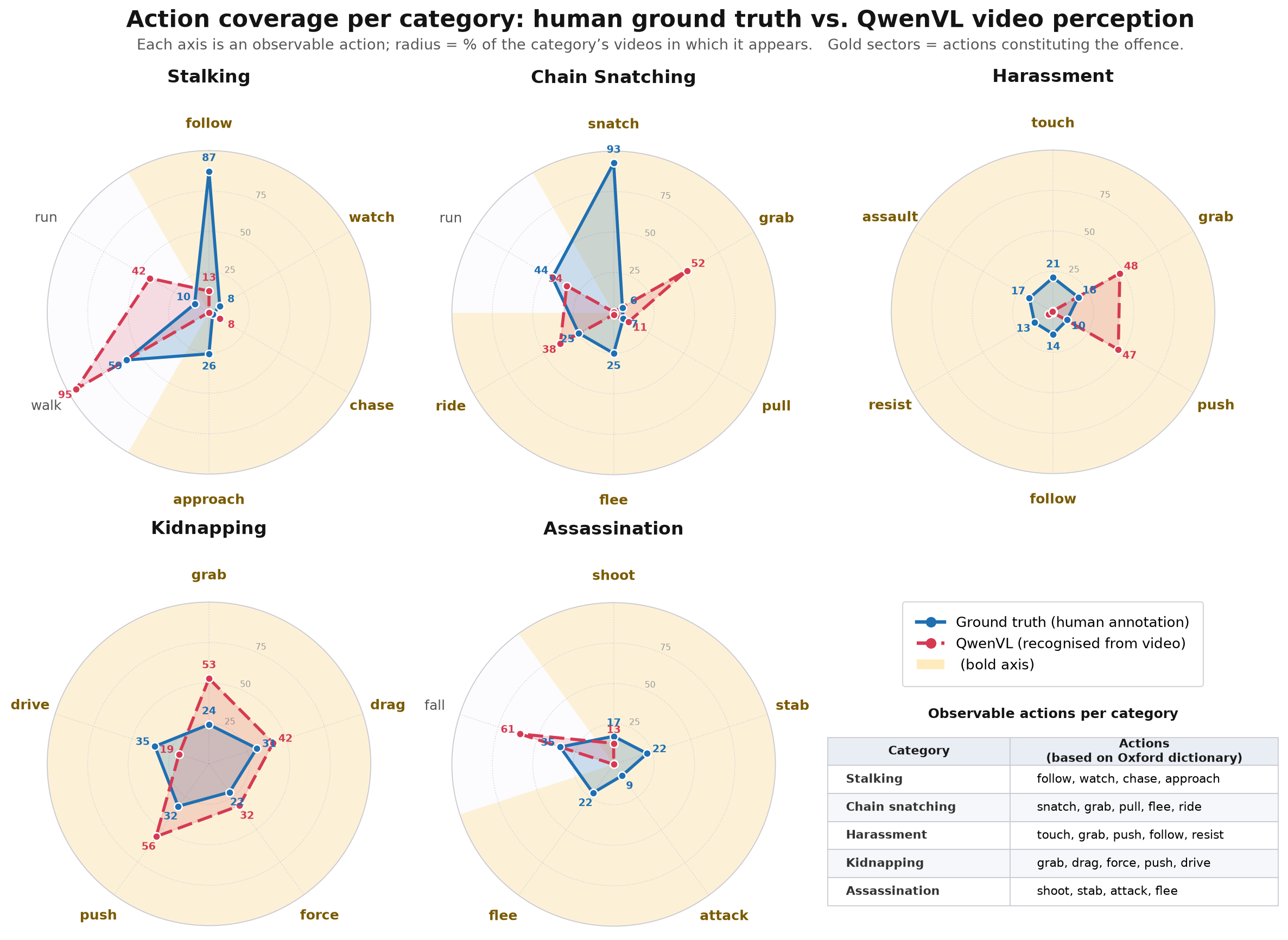}
    \caption{Category-wise comparison of action coverage between human-annotated ground-truth descriptions and QwenVL-generated descriptions. Gold sectors denote crime-constituting actions based on the Oxford Dictionary definition of the offence. Human annotations consistently emphasize these defining actions demonstrating the richer semantic supervision provided by ExtrAnom.} 
    \label{fig:impactions}
\end{figure}

\subsection{Fine-grained Action Annotations}

Existing anomaly benchmarks primarily focus on coarse anomaly categories, whereas subtle human interactions characterize women-centric anomalies. Therefore, fine-grained action annotations are essential for accurately representing these events. Figure~\ref{fig:impactions} compares the category-wise action coverage of human-annotated descriptions and QwenVL~\cite{qwen} generated descriptions. The highlighted sectors denote the crime-constituting actions associated with each anomaly category. Human annotations consistently capture these defining actions, such as \emph{follow} for stalking, \emph{snatch} for chain snatching, \emph{touch} for harassment, \emph{drag} and \emph{drive} for kidnapping, and \emph{shoot} and \emph{stab} for assassination. In contrast, QwenVL assigns greater importance to generic actions such as \emph{walk}, \emph{grab}, \emph{push}, and \emph{fall}, while providing limited coverage of the actions that constitute the offense. This mismatch reduces the semantic distinctiveness between anomaly categories. These results demonstrate that ExtrAnom provides fine-grained, human-curated annotations that preserve category-specific action semantics, making it a more reliable benchmark for developing and evaluating women-centric anomaly detection models.

\section{Conclusion and Future Work}
\label{sec:conclusion}
In this paper, we release ExtrAnom, a women-centric, low-light, and low-resolution video dataset for VAD tasks. The dataset includes different categories of women-centric anomalies, along with human and LLM-generated video-level descriptions. We have benchmarked the dataset on recently proposed multi-modal LLMs and fine-tuned the latest VLM. We have compared video-level descriptions with ground truths using well-known metrics. Our findings reveal that the multi-modal LLMs lack in generating accurate video-level descriptions of women-centric crimes. Next, we have validated the dataset on SOTA vision-based models. The results highlight the importance of a women-centric dataset. We believe it will promote future research on robust, explainable, and multimodal video anomaly understanding. Future work involves the development of focused VLMs that can accurately generate video-level descriptions of such events. Moreover, it will be interesting to observe if the dataset performs equally well on anomalies involving men.

\bibliographystyle{plainnat}
\bibliography{references}

\end{document}